\documentclass[sigconf, authorversion]{acmart}

\usepackage{multirow} 
\usepackage{caption}
\usepackage{subcaption}
\usepackage{amsmath}

\AtBeginDocument{%
  \providecommand\BibTeX{{%
    \normalfont B\kern-0.5em{\scshape i\kern-0.25em b}\kern-0.8em\TeX}}}

\copyrightyear{2021} 
\acmYear{2021} 
\setcopyright{acmlicensed}
\acmConference[C\&C '21]{Creativity and Cognition}{June 22--23, 2021}{Virtual Event, Italy}
\acmBooktitle{Creativity and Cognition (C\&C '21), June 22--23, 2021, Virtual Event, Italy}
\acmPrice{15.00}
\acmDOI{10.1145/3450741.3465263}
\acmISBN{978-1-4503-8376-9/21/06}

\begin{document}

\title{Empirically Evaluating Creative Arc Negotiation for Improvisational Decision-making}

\author{Mikhail Jacob}
\authornote{Work done while at the Georgia Institute of Technology.}
\email{t-mijaco@microsoft.com}
\orcid{1234-5678-9012}
\affiliation{
  \institution{Microsoft Research}
  \city{Cambridge}
  \country{UK}
}

\author{Brian Magerko}
\email{magerko@gatech.edu}
\affiliation{
  \institution{Georgia Institute of Technology}
  \streetaddress{85 5th Street}
  \city{Atlanta}
  \country{USA}
}

\renewcommand{\shortauthors}{Jacob and Magerko}

\begin{abstract}
    Action selection from many options with few constraints is crucial for improvisation and co-creativity. Our previous work proposed creative arc negotiation to solve this problem, i.e., selecting actions to follow an author-defined `creative arc' or trajectory over estimates of novelty, unexpectedness, and quality for potential actions. The CARNIVAL agent architecture demonstrated this approach for playing the Props game from improv theatre in the Robot Improv Circus installation. This article evaluates the creative arc negotiation experience with CARNIVAL through two crowdsourced observer studies and one improviser laboratory study. The studies focus on subjects' ability to identify creative arcs in performance and their preference for creative arc negotiation compared to a random selection baseline. Our results show empirically that observers successfully identified creative arcs in performances. Both groups also preferred creative arc negotiation in agent creativity and logical coherence, while observers enjoyed it more too.
\end{abstract}



\maketitle

\section{Introduction}\label{sec:intro}

Improvisation is a ubiquitous activity that humans do every day. Professional improvisers perform to crowds, regularly finding the perfect response just in time for the unfolding action on stage. While experienced improv performers make the task look easy, improvisational performance is supremely challenging for many reasons \cite{pressing1984cognitive,johnson-laird_how_2002}. Research in human-computer improvisation enables us to create novel forms of artistic expression and interactive experience that combine the social creativity and engagement of improvisation with the scale and democratisation of artificial intelligence (AI). However, improvising with humans poses additional challenges for any improvisational AI agent \cite{long2020whydontcomputersimprovise}.
 
One of the primary challenges for improvisational agents that need to improvise with humans outside of small-scale research prototypes is choosing their next action from a vast set of options in near real-time without a small, well-defined set of goals or hard constraints to optimise. This is called the \textit{improvisational action selection} problem \cite{jacob2019improvisational}. Failure to address this problem can result in incoherent behavior, decision paralysis, or repetitive responses \cite{jacob2019improvisational}.

Previous work \cite{jacob2018creative} proposed \textit{creative arc negotiation} as one solution for improvisational action selection inspired by perceived experiential arcs found in many creative fields. An agent using this approach is intrinsically motivated to negotiate an author-specified `creative arc' along with a human improviser, selecting actions to best follow it during an improvised performance (see section \ref{sec:CARNIVAL}). Creative arcs are authored trajectories through a 'creative space', i.e., a three-dimensional space consisting of novelty, unexpectedness, and quality estimates of potential actions. This work adapts \citeauthor{boden_creative_2004}'s \cite{boden_creative_2004} definition of creativity, paraphrased as the novelty (originality, either personally or historically), surprise (violation of expectations), and value (subjective importance or desirability) of a creative artifact. Unexpectedness and quality estimates were used as proxies for the overloaded concepts of surprise and value.

\begin{figure*}[t]
    \centering
     \begin{subfigure}[b]{0.48\textwidth}
         \centering
         \includegraphics[width=\textwidth]{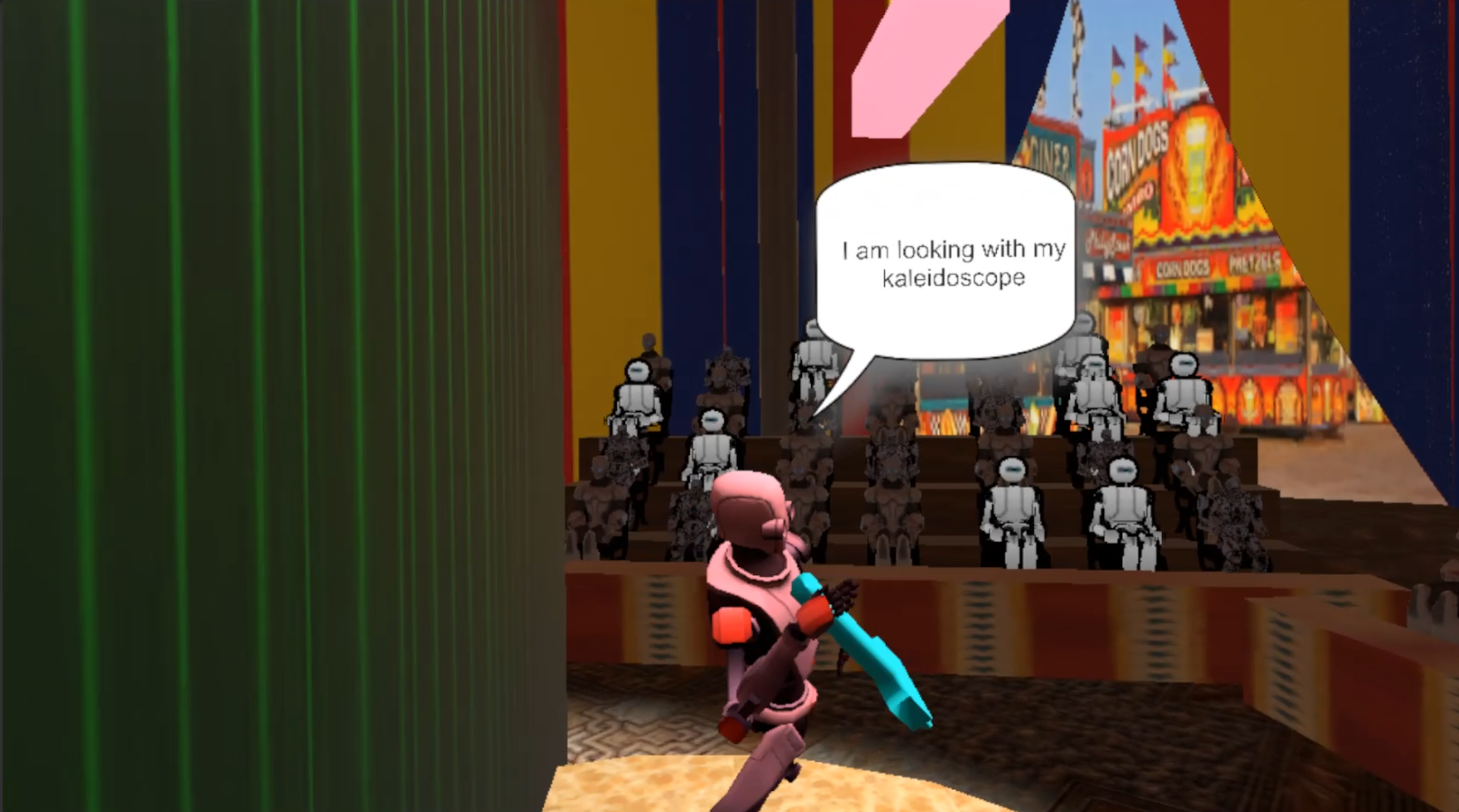}
         \caption{First-person view of agent miming an action with a prop.}
         \label{fig:agent_mime}
     \end{subfigure}
     \hfill
     \begin{subfigure}[b]{0.48\textwidth}
         \centering
         \includegraphics[width=\textwidth]{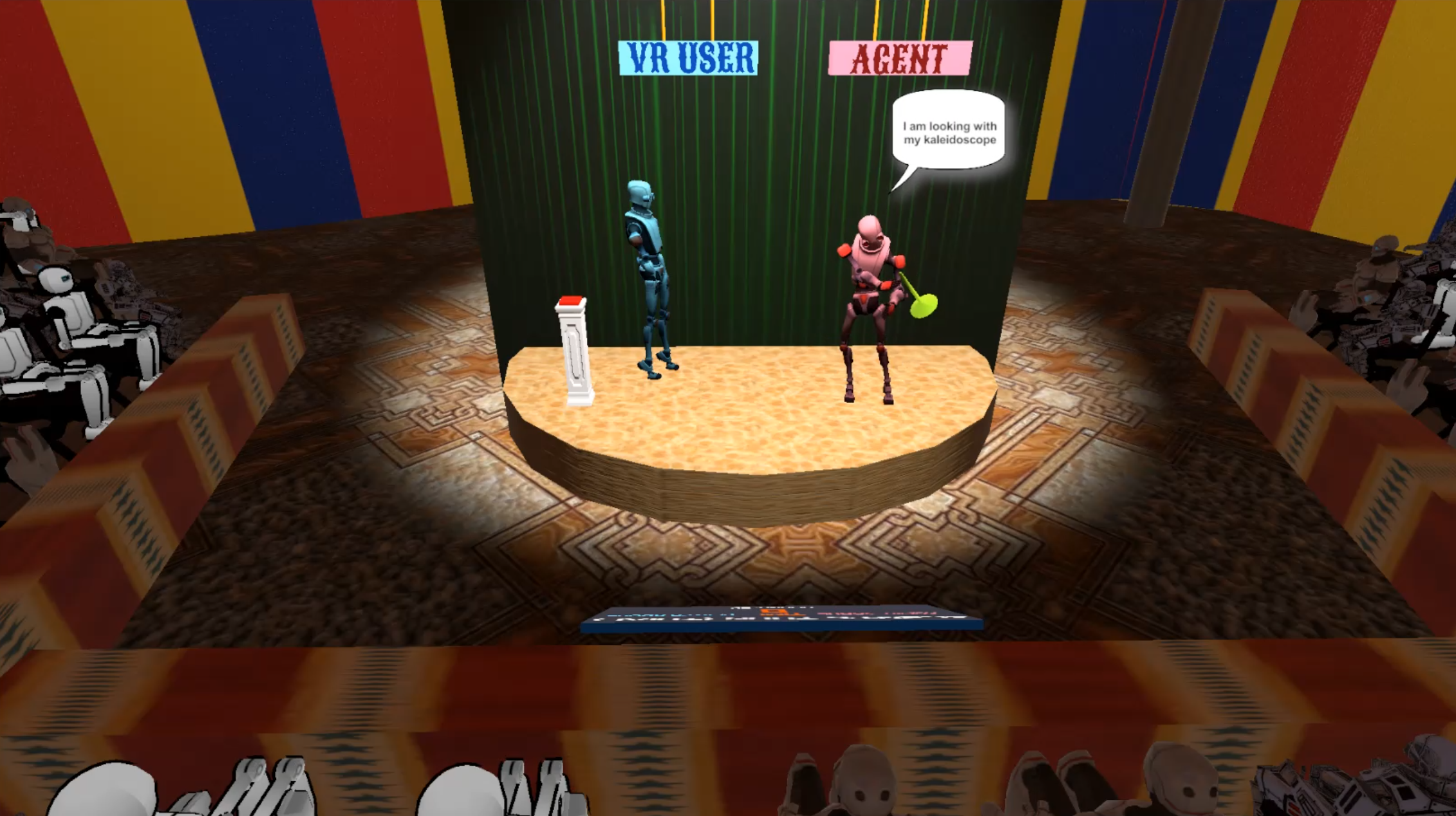}
        \caption{Audience view shown to spectators outside installation.}
        \label{fig:audience_view}
     \end{subfigure}
     \caption{The Robot Improv Circus VR installation. Speech bubble reads, "I am looking with my kaleidoscope."}
     \Description[Two images of the Robot Improv Circus VR installation]{Two images of the Robot Improv circus VR installation. First image shows a first-person view of a robot miming looking through a kaleidoscope with a speech bubble saying as much. Second image shows an audience view of human and virtual agent at that time.}
    \label{fig:RIC_views}
\end{figure*}

We implemented creative arc negotiation in the CARNIVAL agent architecture (section \ref{sec:CARNIVAL}) and studied it within the virtual reality (VR) installation called the Robot Improv Circus \cite{jacob2018creative,jacob2019affordancegeneration}. The installation enables participating improvisers to play an improv game with an AI agent in VR, while an audience views the performance from just outside (section \ref{sec:RIC}). Participants play the Props game from improv theatre in the installation, where an improviser takes turns with a virtual agent to perform actions and dialogue with an abstract prop, pretending it to be a real or fictional object for comedic effect.

This article presents three studies that evaluate the effect of creative arc negotiation on the experience of participating improvisers and observers. We use two large-scale crowdsourced studies of observers and one smaller laboratory study with improvisers to evaluate the following questions. 1) Can observers and improvisers identify creative arcs when an agent uses them for action selection? 2) Do observers and improvisers prefer creative arc negotiation to a random selection baseline in terms of enjoyment, agent creativity, and logical coherence? We present our results and discuss their implications for human-AI improvisation.

\section{Related Work} \label{sec:rel8d-work}


Improvisers demonstrate near real-time collaborative creativity in open-ended and poorly-defined problem domains \cite{johnson-laird_how_2002, mendonca_cognition_2004}. A limited number of improvisational agents exist for theatre \cite{oneill_knowledge-based_2011,brisson2011computational,piplica_full-body_2012,mathewson2017improvised} and storytelling \cite{martin2016improvisational}. The Three Line Scene \cite{oneill_knowledge-based_2011}, Party Quirks \cite{piplica_full-body_2012}, and Tilt Riders \cite{brisson2011computational} systems are cognitive models of improvisational process, using small amounts of hand-authored expert knowledge to improvise. These systems don't explicitly address the improvisational action selection problem when reasoning over larger amounts of knowledge unlike the use of creative arc negotiation in CARNIVAL. More recent improvisational systems such as \cite{mathewson2017improvised,martin2016improvisational} offer exciting solutions for improvisational action selection using larger data sets and machine learning. In addition to addressing this same problem, creative arc negotiation also enables the system to evaluate both the human's and its own actions in the moment while responding, allowing for more creative responsibility and autonomy.


Creative arc negotiation can also be considered a form of intrinsic motivation for agents to follow a given creative arc. This is similar to drives for curiosity (seeking out novel or unexpected stimuli) \cite{schmidhuber2006developmental,merrick2009motivated} or empowerment maximisation (seeking to maximise influence over future outcomes) \cite{salge2016does}. Similarly, evolutionary computing has demonstrated agents trained using novelty search \cite{lehman2011abandoning} and surprise search \cite{gravina2016surprise}, where agents are selected for achieving the most novel or unexpected outcomes instead of the highest quality outcome. Creative arc negotiation operates at a meta level to generalise these approaches beyond pure maximisation, selecting novelty, unexpectedness, and quality to follow a designer-specified creative arc instead. In principle, one could also add other motivations to the negotiation process, for example, to follow an arc of empowerment. Additionally, improvisation is not intended to generate a final product or outcome but on the experiential journey of an ephemeral performance. Therefore, motivating the agent to maximise any aspect of the experience would be counterproductive.


Creative arc negotiation has roots in experience management \cite{bates1992virtual,roberts2008survey} and other interactive narrative research \cite{riedl2013interactive} like  Fa\c{c}ade  \cite{mateas2003faccade} and Merchant of Venice \cite{porteous2011visual}. Fa\c{c}ade sequences hand-authored story fragments or `beats' for the player to follow an arc in dramatic tension. In principle, creative arc negotiation could also emulate dramatic tension search by adding dramatic tension as an added dimension to the negotiation process, while also searching through novelty, unexpectedness, and quality. The Merchant of Venice research \cite{porteous2011visual} describes a visual programming method for drawing dramatic arcs in order to guide a planning-based interactive narrative experience of the Merchant of Venice. This is a potentially useful idea that could be incorporated into CARNIVAL in the future to enable personalization of creative arcs between players or ease the creative arc authoring process for non-experts.

\begin{figure}[h]
    \centering
    \includegraphics[width=\linewidth]{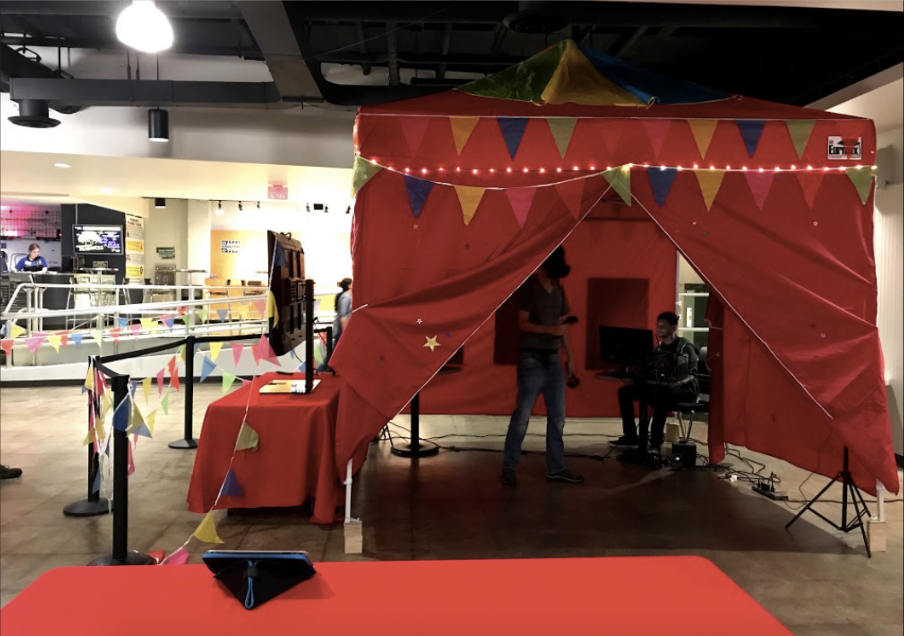}
    \caption{A participant performing in the installation.}
    \Description[The Robot Improv Circus Installation]{A participant is seen miming an action in the Robot Improv Circus VR installation. The installation is housed within a brightly coloured circus tent.}
    \label{fig:RIC_physical_installation}
\end{figure}

\begin{figure*}[t]
    \centering
     \begin{subfigure}[b]{0.50\textwidth}
         \centering
         \includegraphics[width=\textwidth]{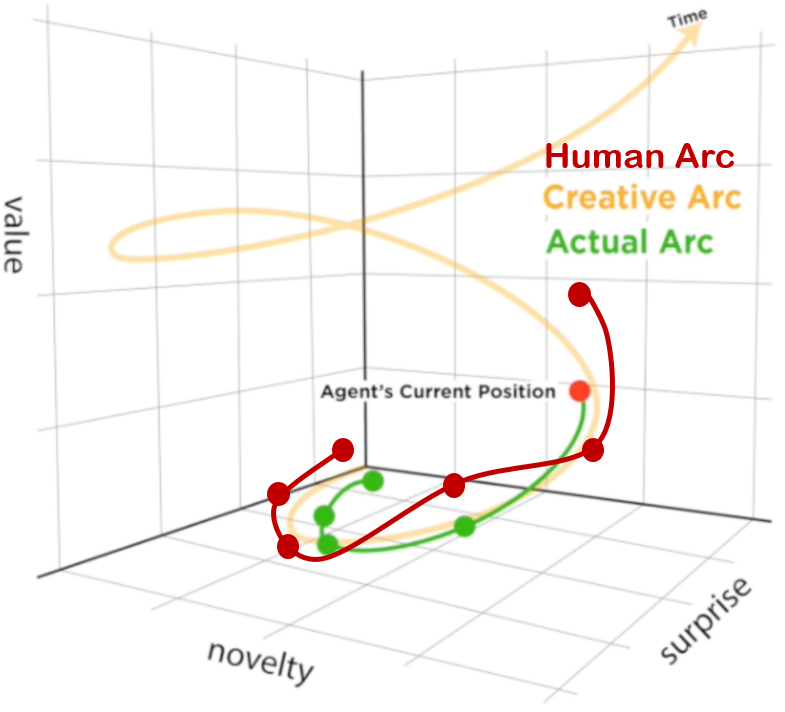}
        \caption{The creative arc negotiation process. The agent follows a designer-given creative arc considering the human's and its own actions in creative space.}
        \label{fig:creative_arcs}
     \end{subfigure}
     \hfill
     \begin{subfigure}[b]{0.46\textwidth}
         \centering
         \includegraphics[width=\textwidth]{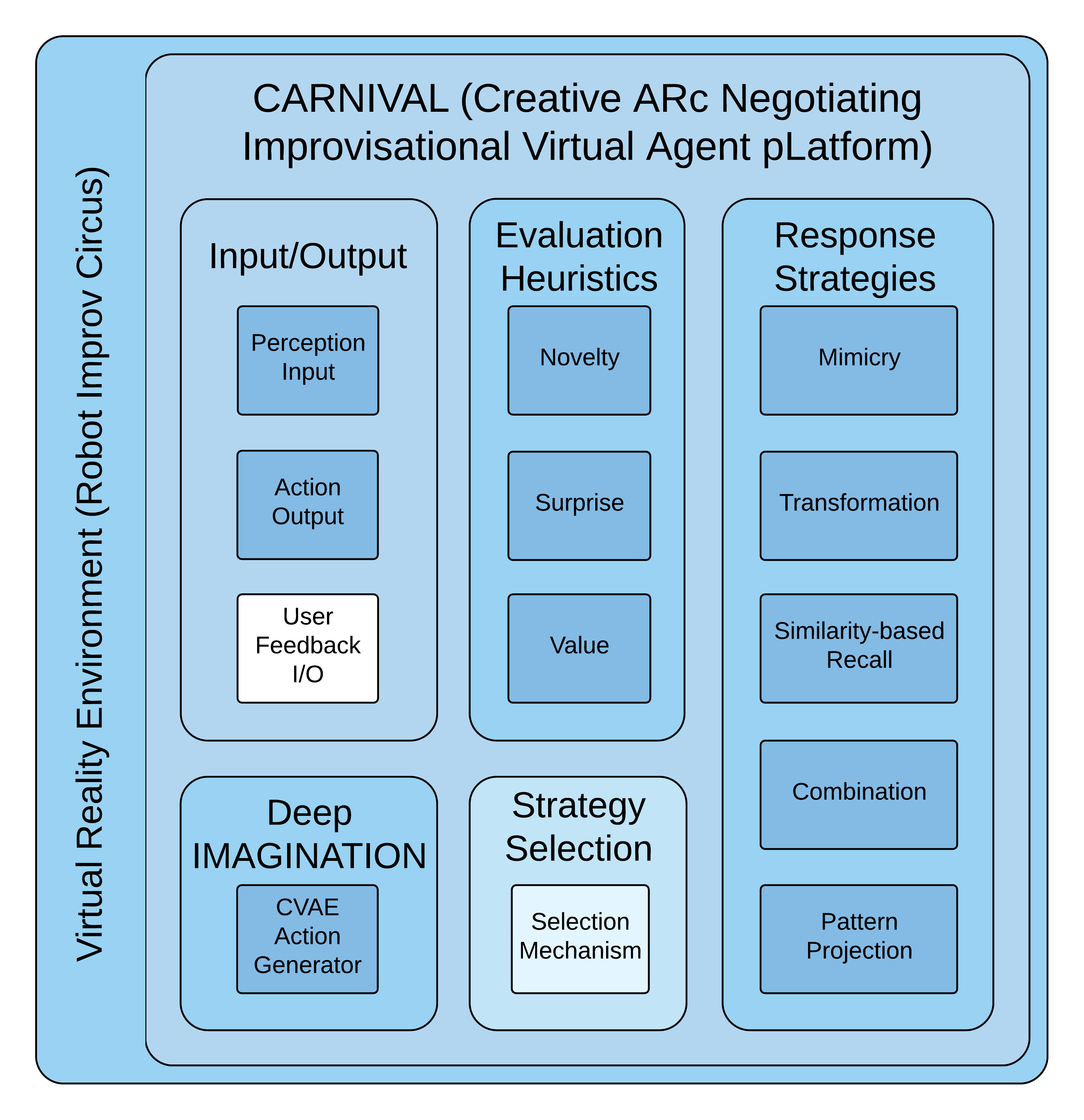}
         \caption{The CARNIVAL agent architecture showing reasoning strategies, action generator, and evaluation heuristics. Partial shading shows future work.}
         \label{fig:CARNIVAL}
     \end{subfigure}
     \caption{Creative arc negotiation as implemented in the CARNIVAL architecture.}
     \Description[Creative arc negotiation in CARNIVAL]{Creative arc negotiation in CARNIVAL is conducted with improvisational reasoning strategies, a deep generative model action generator, and creativity evaluation heuristics.}
    \label{fig:creative_arc_negotiation_in_CARNIVAL}
\end{figure*}

\section{The Robot Improv Circus} \label{sec:RIC}


Improvisers usually play the Props game by taking short turns to perform quick, one-shot actions and dialogue with an abstract prop, pretending it to be a real or fictional object for comedic effect. The Props game allowed us to study short sequences of related actions between players with a common prop, without requiring them to create full narratives. A simplified version of the Props game with a limited set of abstract props and no dialogue exchange was used to study improvisational action selection in the VR installation --- the Robot Improv Circus (figure \ref{fig:RIC_views}) \cite{jacob2018creative}.


The Robot Improv Circus allows players to improvise with a virtual agent. The player, their virtual stage partner, and the audience are all humanoid robots limiting participant expectations for realism. During their turn, the player is given a prop and mimes a pretend action with it. After completing their action, they hit a buzzer to pass their prop to their agent partner. The agent improvises actions with the same prop before hitting the buzzer and passing it back to the player. The agent announces what it is trying to do using text-to-speech audio and a speech bubble. In Fig. \ref{fig:RIC_views}, the agent is trying to mime looking through a kaleidoscope with a long thin prop. The game ends after a fixed number of turns. The improvisation is showcased to external spectators outside the installation using a virtual audience view.

\section{Creative Arc Negotiation in CARNIVAL} \label{sec:CARNIVAL}

Creative arc negotiation for playing the Props game in the Robot Improv Circus is implemented in the CARNIVAL (Creative ARc Negotiating Improvisational Virtual Agent pLatform) agent architecture \cite{jacob2018creative}. The term `negotiation' is used because the agent considers and weights estimates of novelty, unexpectedness, and quality for both the human participant's actions and its own while following a given creative arc. To explain how this process works, we use a running example where the agent has been given a rising creative arc, with novelty, unexpectedness, and quality targets rising over time, and a long, thin, cylindrical prop. CARNIVAL then does the following process. First, given a creative arc to follow (the rising arc), the agent uses a set of improvisational reasoning strategies adapted from previous literature \cite{jacob_interaction-based_2015} to connect its actions to what the human did before and guide its search over potential actions (actions for the long, thin prop). These strategies include \textit{mimicry} (imitating a human action), \textit{transformation} (transforming various elements of a human action before performance), \textit{combination} (combining multiple human actions together), \textit{similarity-based recall} (recalling the nearest or farthest action to a recent human action from episodic memory), and \textit{pattern projection} (looking for temporal patterns between agent-human action pairs and applying that to future human actions).

All reasoning strategies are executed in parallel to generate multiple candidate actions. Actions are generated using a deep generative model called DeepIMAGINATION (Deep IMprovised Action Generation through INteractive Affordance-based exploratION) \cite{jacob2019affordancegeneration}. DeepIMAGINATION is a conditional variational autoencoder (CVAE) \cite{sohn2015learning} that generates mimed gestures conditioned on the physical affordances \cite{gibson1979ecological} of the agent's current prop (different actions are generated for a long, thin, cylinder vs. a short, thick, pyramid). During training, a data set of high-dimensional human gestures captured in VR and attributes of props used to enact them are fed to the CVAE. The model learns to map gestures to points within a low-dimensional latent space (it learns to group similar gestures together for props with similar physical attributes). During improvisation, the agent can sample from the model's latent space to generate potential actions to consider, correctly conditioned on the physical affordances of the agent's current prop (`stabbing' is generated for a long, thin, cylinder instead of `sipping tea' for a short, thick, pyramid). The reasoning strategies mentioned previously, systematically sample from the model's latent space to generate candidate actions.


Each candidate generated action is then evaluated by heuristics that measure its \textit{novelty}, \textit{unexpectedness}, and \textit{quality} to locate that action in the agent's creative space. So using a long, thin, cylinder as a sword might be low in novelty and surprise, but using it as your unicorn horn might be higher in both. Human actions are also evaluated similarly for the agent to reason about. The evaluation models act on the gestural (the representation of agent movement, say when swinging a `sword') and semantic (natural language labels for pretend actions and objects, like the labels `swing' and `sword') contents of the perceived action. Novelty is measured by the agent as the aggregated distance between a perceived action and other comparable experiences that the agent experienced (the difference between swinging a sword and similar swinging actions). The agent measures unexpectedness (as a proxy for surprise) using a combination of Bayesian Surprise \cite{itti2009bayesian} and direct computation of deviation from expectation \cite{maher_evaluating_2010}. Unexpectedness is computed for both the choice of pretend object and pretend action (how unexpected is the choice of `sword' given a long, thin, cylinder and the choice of `swinging' given a pretend sword). These properties are termed \textit{object surprise} and \textit{action surprise} respectively. Quality (as a proxy for value) is then computed using smoothness of motion (how smoothly a sword was swung) and `recognisability' of the gestural and semantic components of the action respectively (how uniquely representative was the chosen action given other potential choices). Finally, each evaluated action and the agent's estimate of the human participant's last action are aggregated to find its effective location in the creative space. The nearest of these actions to the next point on the target creative arc (so for a rising arc, the next action that is more novel, more unexpected, and higher quality) is chosen for performance and played back.

\section{Methodology}\label{sec:method}

We used the following research questions to evaluate the effect of creative arc negotiation for action selection on user experience in the Robot Improv Circus.

\begin{enumerate}
    \item[RQ1:] Can observers and improvisers identify a creative arc when an agent used it for decision-making?
    \item[RQ2:] Does creative arc negotiation result in better observer and improviser experiences compared to random selection?
\end{enumerate}

RQ1 addresses whether the creative arc used by the agent is evident in the improvised performance regardless of whether subjects enjoyed the performance. Recognition of the different creative arcs would validate the agent's ability to create qualitatively different experiences mirroring the arcs. RQ2 investigates whether subjects prefer improvisation with creative arc negotiation to random action selection in terms of enjoyment, creativity, and logical coherence.

\subsection{Observer Recognition of Creative Arcs (RQ1)} \label{sec:method-observer-rec}


We started our evaluation of RQ1 with a survey-driven, observer-rating study. Observers were considered separately from improvisers due to their crucial but differing role in improvised performing arts like improv theatre.
A hundred non-expert raters on Amazon Mechanical Turk \cite{sheehan2018crowdsourcing} were each asked to watch videos of three different improvised sessions between a researcher and the agent in the Robot Improv Circus installation (see Fig. \ref{fig:audience_view}). For each video, they were then asked questions to choose whether a specific property of the performance was rising, falling, or level. In this configuration, random guessing would have a one in three ($33.33\%$) chance of being correct. The alternative hypotheses ($H_A$) stated that for each evaluated property and a given creative arc, the proportion of subjects identifying the correct arc would differ significantly from those incorrectly identifying other arcs. The null hypotheses ($H_0$) stated that for each evaluated property, there would be no significant differences.

Three creative arcs were used in the recorded performances for comparison --- \textit{rising}, \textit{falling}, and \textit{level} arcs. Values for novelty, unexpectedness, and quality (as defined in section \ref{sec:CARNIVAL}) along these arcs, increased uniformly, decreased uniformly, and stayed the same respectively. For a rising arc, this meant that the agent would try to increase the novelty, unexpectedness, and quality of its actions throughout the performance.

The agent used one of the three creative arcs described above in each video. Subjects were asked to determine whether a rising, falling, or level arc was being used by the agent in that video in terms of the \textit{novelty}, \textit{object surprise}, \textit{action surprise}, \textit{quality}, and \textit{user-defined creativity} of the performance (defined previously in section \ref{sec:CARNIVAL}). All subjects were given the definition of each property alongside the questions except for user-defined creativity. Subjects were asked to define creativity before the rating task started and were reminded to use that definition whenever user-defined creativity was evaluated.

\subsection{Effect of Creative Arc Negotiation on Observer Experience (RQ2)} \label{sec:method-observer-comp}


We used another survey-driven, observer-rating study to evaluate RQ2. The creative arc negotiation agent used either \textit{rising}, \textit{falling}, or \textit{level} creative arcs, exactly as in the previous study (section \ref{sec:method-observer-rec}). Our baseline sampled actions uniformly at random from the agent's latent space (section \ref{sec:CARNIVAL}), ensuring meaningfully generated random actions that were still appropriate for the given prop. We chose this baseline due to a lack of other established action selection mechanisms implied by the improvisational action selection problem.

Subjects for the study consisted of 100 non-expert raters on Amazon Mechanical Turk \cite{sheehan2018crowdsourcing}. They were asked to watch videos of two different sessions between a researcher and the agent using either creative arc negotiation or randomly sampled actions to improvise in the Robot Improv Circus. For each video, they were then asked to choose whether they preferred the one on the left or the one on the right in a forced-choice configuration based on different perceived properties of the performance. In this configuration, each video had a random probability of being selected $50\%$ of the time. The different qualities they were asked to compare were \textit{enjoyment}, \textit{user-defined creativity of the agent}, and \textit{logical coherence}. At the start of the study, participants were made to define creativity and reminded to use that definition during the task. 

The initial experiment was also repeated with an identical methodology using videos with just the agent's turns spliced together from the original performance videos (the researcher's actions were removed). This was done to mitigate any potential bias in the results emerging from the human's actions, i.e., in case their actions contributed positively or negatively to observer preferences.
The sample size was also increased to 120 participants.

\subsection{Creative Arc Negotiation and Improviser Experience (RQ1 + RQ2)}\label{sec:method-lab-study}



We conducted an improviser-rating, laboratory study with non-experts to get quantitative and qualitative feedback about the improviser experience of interacting with an agent using creative arc negotiation. The in-person experiment combined our methodologies from the two observer experiments and
asked improvisers to identify creative arcs (RQ1) and compare experiences between creative arcs or random selection (RQ2).

Eighteen participants were recruited for the initial study in two batches (six and twelve subjects) from a non-expert student population. The number of responses obtained per question were either twelve or eighteen, since additional questions were asked of the second batch of participants. No other differences in methodology existed between these two populations, and the number of responses for each question is noted when reporting results.

Participants were first given
an opportunity to get familiar with how to use the VR system and the specific installation through a tutorial and a set of trial rounds for the installation. Participants were next placed into one of 3 groups at random and continued on to complete two study tasks. Finally, the study concluded after participants were debriefed and compensated for their participation.

\textit{The first experimental task was a comparison between creative arc negotiation and random action selection (RQ2).} Participants were assigned to 3 groups. Each group had 1 of 3 possible creative arcs and 1 no arc session.
The ordering for conditions within each group was randomized across participants. Each participant was asked to perform two sessions of improvisation with the agent. In each session, the agent used a different action selection mechanism according to the participant's assigned group. After improvising with the agent twice, the participant was asked to compare the two sessions through a survey followed by a semi-structured interview.

The session comparison questionnaire for these tasks asked the following two to three questions (depending on the study batch). 
\begin{enumerate}
    \item Which of the sessions did you enjoy more?
    \item In which of the sessions would you say your partner was more creative overall?
    \item Which of the two sessions would you say was more logical overall?
\end{enumerate}
These questions received 18, 18, and 12 responses.
For the second question, participants were asked to reflect on their own definition of creativity before completing this questionnaire.
For all questions, participants could select between the options --- session one, session two, both equally, and neither. During the semi-structured interview, participants were asked questions to clarify their definition of creativity used in the questionnaire, memorable reasons or examples of interactions that led to their responses, and other reasons why they preferred one session over the other. Participants were also asked for open-ended feedback on the interaction, experience, or any other aspect of the sessions.

\textit{The second experimental task was creative arc recognition (RQ1).} Participants were assigned to 3 groups, 1 for each pairing of rising, falling, and level creative arcs.
The ordering of conditions was randomised within each group. Participants performed two sessions of improvisation with the agent, answering questions after each session.

Each session was evaluated with a questionnaire. It asked participants whether the novelty, object surprise, action surprise, and user-defined creativity increased, decreased, or stayed the same over time. 
Definitions for each property (except user-defined creativity) were presented to them alongside each question as defined in section \ref{sec:CARNIVAL}. 

\begin{table*}[h]
    \begin{center}
    \caption{Relative recognition percentages between arc types in creative arc identification task (RQ1). Bold is higher between pairs.}
    \label{tab:rfl-tnosasqc}
        \begin{tabular}{lcccccc}
            \toprule
                \multirow{2}{*}{} &
                \multicolumn{2}{c}{\textbf{Rising}} &
                \multicolumn{2}{c}{\textbf{Falling}} &
                \multicolumn{2}{c}{\textbf{Level}} \cr
                & {\textbf{Correct}} & {\textbf{Incorrect}} & {\textbf{Correct}} & {\textbf{Incorrect}} & {\textbf{Correct}} & {\textbf{Incorrect}}\\
                \midrule
                \textbf{Novelty} & $\mathbf{57.14\%}$ & $42.86\%$ & $37.14\%$ & $\mathbf{62.86\%}$ & $20.95\%$ & $\mathbf{79.05\%}$ \\
                \textbf{Object Surprise} & $\mathbf{53.33\%}$ & $46.67\%$ & $44.76\%$ & $\mathbf{55.24\%}$ & $34.29\%$ & $\mathbf{65.71\%}$ \\
                \textbf{Action Surprise} & $47.12\%$ & $\mathbf{52.88\%}$ & $37.14\%$ & $\mathbf{62.86\%}$ & $42.86\%$ & $\mathbf{57.14\%}$ \\
                \textbf{Quality} & $\mathbf{73.08\%}$ & $26.92\%$ & $\mathbf{61.90\%}$ & $38.10\%$ & $\mathbf{60.00\%}$ & $40.00\%$ \\
                \textbf{Creativity} & $\mathbf{51.43\%}$ & $48.57\%$ & $43.81\%$ & $\mathbf{56.19\%}$ & $33.33\%$ & $\mathbf{66.67\%}$ \\
            \bottomrule
        \end{tabular}
    \end{center}
\end{table*}

\begin{table*}[h]
    \begin{center}
    \caption{Chi-square goodness of fit for creative arc identification task arcs (RQ1). Bold significant at $p<0.05$. $\phi_c$ is effect size.}
    \label{tab:csgf-tnosasqc}
        \begin{tabular}{lccccccccc}
            \toprule
                \multirow{2}{*}{} &
                \multicolumn{3}{c}{\textbf{Rising}} &
                \multicolumn{3}{c}{\textbf{Falling}} &
                \multicolumn{3}{c}{\textbf{Level}} \cr
                & {$\boldsymbol{\chi^2}$} & {\textbf{p}} & {$\boldsymbol{\phi_c}$} & {$\boldsymbol{\chi^2}$} & {\textbf{p}} & {$\boldsymbol{\phi_c}$} & {$\boldsymbol{\chi^2}$} & {\textbf{p}} & {$\boldsymbol{\phi_c}$} \\
                \midrule
                \textbf{Novelty} & $\mathbf{26.79}$ & $\mathbf{<10^{-5}}$ & $\mathbf{0.505}$ & $0.69$ & $0.40763$ & $0.081$  & $\mathbf{7.24}$ & $\mathbf{0.0071}$ & $\mathbf{0.263}$  \\
                \textbf{Object Surprise} & $\mathbf{18.90}$ & $\mathbf{<10^{-5}}$ & $\mathbf{0.424}$ & $\mathbf{6.17}$ & $\mathbf{0.013}$ & $\mathbf{0.242}$  & $0.04$ & $0.836$ & $0.020$  \\
                \textbf{Action Surprise} & $\mathbf{8.89}$ & $\mathbf{0.0029}$ & $\mathbf{0.292}$ & $0.69$ & $0.4076$ & $0.081$  & $\mathbf{4.29}$ & $\mathbf{0.0384}$ & $\mathbf{0.202}$  \\
                \textbf{Quality} & $\mathbf{73.92}$ & $\mathbf{<10^{-5}}$ & $\mathbf{0.843}$ & $\mathbf{38.57}$ & $\mathbf{<10^{-5}}$ & $\mathbf{0.606}$  & $\mathbf{33.60}$ & $\mathbf{<10^{-5}}$ & $\mathbf{0.566}$  \\
                \textbf{Creativity} & $\mathbf{15.47}$ & $\mathbf{0.00008}$ & $\mathbf{0.384}$ & $\mathbf{5.19}$ & $\mathbf{0.0228}$ & $\mathbf{0.222}$  & $0$ & $1$ & $0$  \\
            \bottomrule
        \end{tabular}
    \end{center}
\end{table*}

\section{Results}\label{sec:result}

We present our results on the effect of creative arc negotiation on observer and improviser experiences in the Robot Improv Circus in this section. We start with the results for each study (observer creative arc recognition, observer creative arc comparison, and participant creative arc recognition with comparison) and then discuss their implications in the next section.

\subsection{Observer Recognition of Creative Arcs (RQ1)} \label{sec:result-obs-identify}


The relative percentages of participants who correctly (and incorrectly) identified each property of the performance as rising, falling, or level are presented in Table \ref{tab:rfl-tnosasqc}. Note that random guessing would score approximately $33.33\%$. The total parameter reports accuracy over all other parameters combined.

A 
chi-square goodness of fit test was performed to evaluate whether there were significant results between correctly vs. incorrectly identifying the direction of the arc for the given property. The null hypotheses 
for the five questions were that there was no significant difference between the distributions of responses identifying the arc for each property. The alternate hypotheses 
stated that significant differences did exist between these distributions. The results for each rated property of the session from the chi-square goodness of fit test are presented in table \ref{tab:csgf-tnosasqc}.

The creative arc recognition accuracies and the statistical hypotheses testing showed that for
\textit{novelty}, \textit{object surprise}, and \textit{creativity}, performances with \textit{rising} and \textit{falling} arcs could be identified reliably. Surprisingly, it also showed that recognition accuracies for all significantly differing properties of \textit{level} arcs were consistently as bad as random guessing. 

The chi-square goodness of fit results showed medium and large effects for a subset of significant results. The results indicate that observers reliably noticed the intended changes in those properties during the performance. Most importantly, these results conclude that \textbf{for observers, most queried properties defining our creative space were reliably identifiable with rising arcs, less reliably identifiable with falling arcs, and difficult to identify for level arcs.} 

\subsection{Effect of Creative Arc Negotiation on Observer Experience (RQ2)}  \label{sec:result-obs-comp}


The percentage of participants who preferred creative arc negotiation over no arc random sampling for each rated property of the performance are presented in table \ref{tab:rfl-teac}. The properties that participants were asked about included which session they enjoyed more (\textit{enjoyment}), in which session the agent seemed more creative (\textit{Agent Creativity}), and which session seemed more logically coherent (\textit{Coherence}). Note that random guessing would score 50\% in this task.

A 
chi-square goodness of fit test was performed on all the queried properties to evaluate whether there were significant differences in observer preferences between creative arc and no arc conditions for each arc. The null hypotheses 
for comparing each queried property were that there were no significant differences between the distributions of responses for each arc to the distribution of expected outcomes in each case. The alternate hypotheses 
stated that significant differences did exist between these distributions.
The results for each rated property of the session from the chi-square goodness of fit test are presented in table \ref{tab:csgf-tecc}.

\begin{table*}[h]
    \begin{center}
    \caption{Relative preferences between `arc' and `no arc' conditions in creative arc comparison task (RQ2). Bold is higher in pair.}
    \label{tab:rfl-teac}
        \begin{tabular}{lcccccc}
            \toprule
                \multirow{2}{*}{} &
                \multicolumn{2}{c}{\textbf{Rising}} &
                \multicolumn{2}{c}{\textbf{Falling}} &
                \multicolumn{2}{c}{\textbf{Level}} \cr
                & {\textbf{Arc}} & {\textbf{No Arc}} & {\textbf{Arc}} & {\textbf{No Arc}} & {\textbf{Arc}} & {\textbf{No Arc}}\\
                \midrule
                \textbf{Enjoyment} & $\textbf{63.46\%}$ & $36.54\%$ & $\textbf{62.50\%}$ & $37.50\%$ & $28.57\%$ & $\textbf{71.43\%}$ \\
                \textbf{Agent Creativity} & $\textbf{64.08\%}$ & $35.92\%$ & $\textbf{56.73\%}$ & $43.27\%$ & $31.73\%$ & $\textbf{68.27\%}$ \\
                \textbf{Coherence} & $\textbf{80.58\%}$ & $19.42\%$ & $\textbf{75.96\%}$ & $24.04\%$ & $24.04\%$ & $\textbf{75.96\%}$ \\
            \bottomrule
        \end{tabular}
    \end{center}
\end{table*}

\begin{table*}[h]
    \begin{center}
    \caption{Chi-square goodness of fit for creative arc comparison task arcs (RQ2). Bold significant at $p<0.05$. $\phi_c$ is effect size.}
    \label{tab:csgf-tecc}
        \begin{tabular}{lccccccccc}
            \toprule
                \multirow{2}{*}{} &
                \multicolumn{3}{c}{Rising} &
                \multicolumn{3}{c}{Falling} &
                \multicolumn{3}{c}{Level} \cr
                & {$\boldsymbol{\chi^2}$} & {\textbf{p}} & {$\boldsymbol{\phi_c}$} & {$\boldsymbol{\chi^2}$} & {\textbf{p}} & {$\boldsymbol{\phi_c}$} & {$\boldsymbol{\chi^2}$} & {\textbf{p}} & {$\boldsymbol{\phi_c}$} \\
                \midrule
                \textbf{Enjoyment} & $\mathbf{7.54}$ & $\mathbf{0.00604}$ & $\mathbf{0.269}$ & $\mathbf{6.5}$ & $\mathbf{0.01079}$ & $\mathbf{0.250}$  & $\mathbf{19.29}$ & $\mathbf{<10^{-5}}$ & $\mathbf{0.429}$  \\
                \textbf{Agent Creativity} & $\mathbf{8.17}$ & $\mathbf{0.00427}$ & $\mathbf{0.282}$ & $\mathbf{1.89}$ & $\mathbf{0.013}$ & $\mathbf{0.135}$  & $\mathbf{13.89}$ & $\mathbf{0.00019}$ & $\mathbf{0.365}$  \\
                \textbf{Coherence} & $\mathbf{38.53}$ & $\mathbf{<10^{-5}}$ & $\mathbf{0.612}$ & $\mathbf{28.04}$ & $\mathbf{<10^{-5}}$ & $\mathbf{0.519}$  & $\mathbf{28.04}$ & $\mathbf{<10^{-5}}$ & $\mathbf{0.519}$  \\
            \bottomrule
        \end{tabular}
    \end{center}
\end{table*}

The results of this experiment suggest that there were \textbf{significant, reliably detectable preferences for \textit{rising} and \textit{falling} creative arc negotiation performances in comparison with a random sampling baseline}, at least for observers viewing videos of performances. \textit{Level} arcs significantly and reliably were preferred less than the random selection baseline. All properties (\textit{enjoyment}, \textit{agent creativity}, and \textit{coherence}) showed significant results with effect sizes ranging from small to large. The effects for rising and falling arcs (with the effect stronger in general for rising arcs) showed that coherence was the most improved, with agent creativity and enjoyment following closely behind.

The results from the repeated study with just the agent's actions spliced together in the observer's video were analyzed exactly the same way as the previous analysis. The results for recognition accuracies across arcs can be seen in table \ref{tab:rfl-oa-teac}. After performing statistical significance testing, the results can be seen in table
\ref{tab:csgf-oa-tecc}.

\begin{table*}[h]
    \begin{center}
    \caption{Relative preferences between an arc condition and a no arc condition in creative arc comparison task (RQ2) with only agent turns (no human turns). Bold is higher between pairs. Surprisingly, preferences are stronger without human turns.}
    \label{tab:rfl-oa-teac}
        \begin{tabular}{lcccccc}
            \toprule
                \multirow{2}{*}{} &
                \multicolumn{2}{c}{\textbf{Rising}} &
                \multicolumn{2}{c}{\textbf{Falling}} &
                \multicolumn{2}{c}{\textbf{Level}} \cr
                & {\textbf{Arc}} & {\textbf{No Arc}} & {\textbf{Arc}} & {\textbf{No Arc}} & {\textbf{Arc}} & {\textbf{No Arc}}\\
                \midrule
                \textbf{Enjoyment} & $\textbf{86.67\%}$ & $13.33\%$ & $\textbf{70.83\%}$ & $29.17\%$ & $23.53\%$ & $\textbf{76.47\%}$ \\
                \textbf{Agent Creativity} & $\textbf{73.33\%}$ & $26.67\%$ & $\textbf{63.03\%}$ & $36.97\%$ & $25.21\%$ & $\textbf{74.79\%}$ \\
                \textbf{Coherence} & $\textbf{93.33\%}$ & $6.67\%$ & $\textbf{73.33\%}$ & $26.67\%$ & $16.81\%$ & $\textbf{83.19\%}$ \\
            \bottomrule
        \end{tabular}
    \end{center}
\end{table*}

\begin{table*}[h]
    \begin{center}
    \caption{Chi-square goodness of fit for creative arc comparison task (RQ2) with only agent turns (no human turns). Bold significant at $p<0.05$. $\phi_c$ is effect size. Results are stronger without human turns.}
    \label{tab:csgf-oa-tecc}
        \begin{tabular}{lccccccccc}
            \toprule
                \multirow{2}{*}{} &
                \multicolumn{3}{c}{\textbf{Rising}} &
                \multicolumn{3}{c}{\textbf{Falling}} &
                \multicolumn{3}{c}{\textbf{Level}} \cr
                & {$\boldsymbol{\chi^2}$} & {\textbf{p}} & {$\boldsymbol{\phi_c}$} & {$\boldsymbol{\chi^2}$} & {\textbf{p}} & {$\boldsymbol{\phi_c}$} & {$\boldsymbol{\chi^2}$} & {\textbf{p}} & {$\boldsymbol{\phi_c}$} \\
                \midrule
                \textbf{Enjoyment} & $\mathbf{64.53}$ & $\mathbf{<10^{-5}}$ & $\mathbf{0.733}$ & $\mathbf{20.83}$ & $\mathbf{<10^{-5}}$ & $\mathbf{0.417}$  & $\mathbf{33.35}$ & $\mathbf{<10^{-5}}$ & $\mathbf{0.529}$  \\
                \textbf{Agent Creativity} & $\mathbf{26.13}$ & $\mathbf{<10^{-5}}$ & $\mathbf{0.467}$ & $\mathbf{8.08}$ & $\mathbf{0.00449}$ & $\mathbf{0.261}$  & $\mathbf{29.25}$ & $\mathbf{<10^{-5}}$ & $\mathbf{0.496}$  \\
                \textbf{Coherence} & $\mathbf{90.13}$ & $\mathbf{<10^{-5}}$ & $\mathbf{0.867}$ & $\mathbf{26.13}$ & $\mathbf{<10^{-5}}$ & $\mathbf{0.467}$  & $\mathbf{52.45}$ & $\mathbf{<10^{-5}}$ & $\mathbf{0.664}$  \\
            \bottomrule
        \end{tabular}
    \end{center}
\end{table*}

The results for the repeated observer study with footage of just the agent taking its turns \textbf{showed an even stronger effect in the same direction as the version with the human performer.} This allowed us to remove the effect of the human on the observed results. It also allowed us to address any potential concerns about researcher bias in terms of implicitly shaping the videos for evaluation. This is a valid concern since it is a co-creative performance with creative responsibilities falling on the shoulders of both human and computer improviser. It would be natural for there to be researcher bias or error while constructing the comparison videos. However, the results from the repeated iteration of the study lay any such concerns to rest and improve on the previous results in terms of effect size and increased preference for the creative arc negotiation versions of the system. 

\subsection{Creative Arc Negotiation and Improviser Experience (RQ1 + RQ2)} \label{sec:result-lab-study}



The results for the creative arc comparison and identification tasks are presented in this section.

\subsubsection{Creative Arc Comparison (RQ2)} \label{sec:result-lab-comp}

The results from the questionnaire for the creative arc comparison study task are summarized and presented in Tables \ref{tab:eac-anabn} and \ref{tab:csgf-eac-anabn}.
Table \ref{tab:eac-anabn} summarizes the relative differences in four possible preferences (creative arc, no arc, both, or neither) between the two conditions compared in the task (creative arc and no arc). Table \ref{tab:csgf-eac-anabn} shows the result of performing a chi-square goodness of fit test on the combined data for comparing creative arc sessions against no arc sessions. The null hypotheses for each queried property compared was that there were no significant differences between the distribution of preferences for creative arc negotiation sessions to the distribution of expected outcomes. The alternate hypotheses for these properties were that there were significant differences for these distributions. For both analyses of the creative arc comparison task, our sample size was too small to split the different comparisons by arc types. Therefore, we did not separately compare preferences for each arc type against no arc sessions.

\begin{table*}[h]
    \begin{center}
    \caption{Relative preferences for 'arc', 'no arc', 'both', or 'neither' between a creative arc session and a 'no arc' (random action selection) session in the participant-rating creative arc comparison task (RQ2). Bold is highest for the row. N is sample size.}
    \label{tab:eac-anabn}
        \begin{tabular}{lccccc}
            \toprule
                & {\textbf{Creative Arc}} & {\textbf{No Arc}} & {\textbf{Both}} & {\textbf{Neither}} & {\textbf{N}} \\
                \midrule
                \textbf{Enjoyment} & $\textbf{38.89\%}$ & $33.33\%$ & $27.78\%$ & $0\%$ & $18$ \\
                \textbf{Agent Creativity} & $\textbf{55.56\%}$ & $27.78\%$ & $5.56\%$ & $11.11\%$ & $18$ \\
                \textbf{Coherence} & $\textbf{58.33\%}$ & $8.33\%$ & $25\%$ & $8.33\%$ & $12$ \\
            \bottomrule
        \end{tabular}
    \end{center}
\end{table*}

\begin{table*}[h]
    \begin{center}
        \caption{Chi-square goodness of fit between 'arc' and 'no arc' sessions for the participant-rating creative arc comparison task (RQ2). Bold significant at $p<0.05$. $\phi_c$ and $\tilde{\phi_c}$ are effect size and adjusted effect size respectively. N is sample size.}
        \label{tab:csgf-eac-anabn}
        \begin{tabular}{lccccc}
            \toprule
                & {$\boldsymbol{\chi^2}$} & {\textbf{p}} & {$\boldsymbol{\phi_c}$} & {$\boldsymbol{\tilde{\phi_c}}$} & {\textbf{N}} \\
                \midrule
                \textbf{Enjoyment} & $6.44$ & $0.09188$ & $0.598$ & $0.439$ & $18$ \\
                \textbf{Agent Creativity} & $\mathbf{10.89}$ & $\mathbf{0.01234}$ & $\mathbf{0.778}$ & $\mathbf{0.675}$ & $\mathbf{18}$ \\
                \textbf{Coherence} & $\mathbf{8}$ & $\mathbf{0.04601}$ & $\mathbf{0.816}$ & $\mathbf{0.658}$ & $\mathbf{12}$ \\
            \bottomrule
        \end{tabular}
    \end{center}
\end{table*}

The semi-structured interview data contained explanations for subject questionnaire choices, definitions for creativity, other memorable aspects of the sessions, and feedback about the experience. This data was transcribed and analysed to surface key themes and trends among the subjects' responses. The results offer additional support for the effectiveness of creative arc negotiation on improviser experience.

Participant explanations for why they enjoyed one session over the other included several distinct themes. Two (of 5) subjects who said they enjoyed both sessions equally did so because of how engaging and immersive they found the physicality of interaction in VR, making it harder to notice any differences in the agent's behaviour. The majority of subjects who said they enjoyed the no arc session more (4 of 6), also said that they enjoyed it more because the prop they received in that session was easier to map to different objects for pretending with. This is encouraging, since their preference was not based on the agent's performance. A third of all subjects (6 of 18) described the novelty of agent actions as their main reason for their response. However, a greater number (and proportion) of subjects who enjoyed the creative arc session more provided this reason compared to no arc-preferring subjects (4 of 7 vs. 2 of 6). This also suggests that greater novelty can be attained when novelty is part of the search criteria than with random selection.

Participants also provided the definitions of creativity they used and reasons for why they judged agents in one session more creative than the other. The vast majority of participants (16 of 17 who gave their definition) considered novelty + surprise as main factors in their definition. Some subjects seemed to combine concepts like novelty and surprise in their responses, so the two concepts are combined together here. Similarly, the vast majority of subjects chose one session over the other in terms of agent creativity because of a perceived difference in diversity or novelty (8 of 10 for creative arc sessions and 4 of 5 for no  arc sessions). Four subjects overall considered it more creative when ideas were generated for "harder" props and 2 of 10 subjects chose the creative arc session as a result.

Subjects attributed the logical coherence of sessions (or the lack thereof) to different reasons. Three (of 7) subjects who chose the creative arc session highlighted a perceived narrative or story-like sequence to the agent's actions exemplified by the quote, ``There was more of a sequence of events. \ldots like a beginning, middle, and end to the story \ldots it felt more story-like.'' Two subjects thought neither agent was coherent with 1 saying, "Both were equally illogical." One (of 7) who chose the creative arc session mentioned how the agent paid attention to their actions.

The statistically significant results from the chi-square analysis for the session comparison responses, large effect sizes, the direction of preferences, and subjects' responses in interviews showed promise for the effect of creative arc negotiation on improviser experience, at least compared to a random action selection baseline. \textbf{Subjects reliably and strongly preferred creative arc negotiation when evaluating agent creativity and logical coherence between the two types of sessions.} User enjoyment did not differ significantly between the two conditions. However, further study is required to rule out an effect on user enjoyment, since 6 of 11 subjects who enjoyed arc negotiation less, described coincidental or task-specific differences unrelated to the agent's behaviour as reasons for their choice.

\subsubsection{Creative Arc Identification (RQ1)} \label{sec:result-lab-identify}

The recognition percentages were combined across all creative arc types for each queried property in the session creative arc identification questionnaire. These results showed performances similar to random guessing (at around 33.33\%). These results were also not significant in a chi-square goodness of fit test.
This is a negative result since improvisers could not differentiate between creative arcs driving agent actions.
However, it is important to note that our experimental design used creative arc identification tasks to understand whether subjects could identify trends in their experience that the agent was attempting to modulate regardless of whether they preferred experiences with that form of action selection. We discuss potential reasons for these results in section \ref{sec:discussion}. These initial results for creative arc identification will also be expanded and reviewed in the future after increasing the sample size to get higher confidence results.


\section{Discussion} \label{sec:discussion}

The evaluation studies in this work aimed to grasp the effect of creative arc negotiation on observer and improviser experience. Could they identify what kind of creative arc an agent used in a performance (RQ1)? Would they prefer improvisation sessions guided by creative arc negotiation over an alternative (RQ2)?

The results from the two observer studies provided strong positive evidence that observers both identified and preferred creative arcs, though this phenomenon was applicable to different effect sizes based on the arc. The result that observers preferred performances guided by creative arc negotiation in terms of enjoyment, agent creativity, and logical coherence, empirically demonstrates that \textbf{creative arc negotiation addresses the improvisational action selection problem for observers of improvised performances.}
The existence of these preferences makes sense given the body of literature on the presence of specific sets of arcs across narratives for dramatic tension and plot \cite{aristotle_butcher_1969,freytag1896freytag,vonnegut_2005} or character affect \cite{reagan2016emotional}. The result that observers had a significant preference for random sampling over level arcs was unexpected and requires more study. Perhaps random selection resulted in more novelty. The laboratory study did indicate that novelty was key for perceptions of agent creativity and subject preferences.

The results from the in-person laboratory study indicated that \textbf{improvisers also preferred creative arc negotiation for agent creativity and coherence though not necessarily for enjoyment.} The interviews did partially explain why enjoyment was evenly distributed but further study is needed. The negative results for creative arc identification warrant deeper review. Evidence \cite{kelso_dramatic_1993,laurel1991computers} suggests that interactors in the midst of an ephemeral interactive experience have trouble keeping track of longer-term effects. \citeauthor{kelso_dramatic_1993} \cite{kelso_dramatic_1993} interpret this positively, since interactive narratives then do not need strong narrative coherence, unlike other forms of narrative, if participants cannot keep track of these longer-term links. However, their finding also implies that demonstrating strong trends in the user's experience based on longer-term effects would be harder in these experiences.

Our empirical evaluation above provides initial evidence that creative arc negotiation can be successfully used for improvisational action selection. Further evaluation could consider longer, non-monotonic creative arcs and their use in performances in the wild. Future work could also investigate creative arcs as a technique for negotiating between authorial intent and agent autonomy -- a general challenge in many co-creative and human-AI domains \cite{riedl2013interactive,jacob2020itsunwieldy}.

\section{Conclusion} \label{sec:conclusion}

This article describes creative arc negotiation in the CARNIVAL architecture as a solution to the improvisational action selection problem. We also briefly describe the Props game in the Robot Improv Circus installation as a domain for studying creative arc negotiation. Finally, we contribute three experiments to understand the effect of creative arc negotiation in the Robot Improv Circus on the experience of observers and participating improvisers.

The results from the creative arc identification and comparison studies for observers showed that, depending on the specific arc, they could identify trends in the performances that corresponded to the creative arc used by the agent. The experiments also showed their preference for action selection using creative arc negotiation over a random sampling baseline in terms of perceived enjoyment, agent creativity, and logical coherence. Observer preferences were even stronger when the human partner's actions were removed from the tasks, and only the agent's actions were evaluated. 

Creative arc identification was not successful in our small-scale laboratory study for participating improvisers, unfortunately, and we discussed why this might be the case. More positively, evidence from the creative arc comparison task showed that, depending on the specific arc, perceptions of agent creativity and logical coherence were significantly higher (enjoyment required more study for conclusive results) for improvised performances with creative arc negotiation compared to our baseline. Therefore, \textbf{creative arc negotiation can be successfully used by improvisational agents for action selection and performances created this way are empirically preferred over our baseline by both observers and participating improvisers.}

\begin{acks}
    This material is based on work supported by a Georgia Institute of Technology Office of the Arts \textit{Creative Curricular Initiatives} grant. The authors would like to thank many former and current members of the Expressive Machinery Lab for their contributions, help, support, and advice on this research.
\end{acks}

\bibliographystyle{ACM-Reference-Format}
\bibliography{massive.bib,massive1.bib,massive2.bib,massive3.bib,massive4.bib,sample-base.bib}

\end{document}